\documentclass{article}
\usepackage{aaai}
\usepackage{amssymb}
\usepackage{amsfonts}
\usepackage{latexsym}
\usepackage{graphicx}

\newcommand{\plang}{\cal L}
\newcommand{\nonmono}{\mid \hspace{-0.6mm} \backsim }

\newcommand{\nonexpt}{\mid \hspace{-0.6mm} \backsim_{\leqslant}}

\newcommand{\nonP}{\mid \hspace{-0.6mm} \backsim_{P}}

\newtheorem{definition}{Definition}

\newcommand{\B}{\mathbf{B}}

\title{Applying Maxi-adjustment to Adaptive Information Filtering Agents}

\author{\small Raymond Lau \and \small Arthur H.M. ter Hofstede\\
\small Cooperative Information Systems Research Centre\\
\small Queensland University of Technology\\
\small GPO Box 2434, Brisbane Qld 4001, Australia\\
\small E-mail:\{raymond, arthur\}@icis.qut.edu.au \And
\small Peter D. Bruza\\
\small Distributed Systems Technology Centre\\
\small The University of Queensland\\
\small Brisbane, Qld 4072, Australia\\
\small E-mail:bruza@dstc.edu.au}

\begin{document}

\nocopyright 
\maketitle

\begin{abstract}
\begin{quote}
Learning and adaptation is a fundamental
property of intelligent agents. In the context of adaptive information
filtering, a filtering agent's beliefs about a user's information needs 
have to be revised regularly with reference to the user's most current
information preferences. This learning and adaptation process is essential
for maintaining the agent's filtering performance.
The AGM belief revision paradigm
provides a rigorous foundation for modelling 
rational and minimal changes to an agent's beliefs. In particular, the
maxi-adjustment method, which follows the AGM rationale of belief change,
offers a sound and robust computational mechanism to develop
adaptive agents so that \emph{learning autonomy} of these agents
can be enhanced. This paper describes how the maxi-adjustment
method is applied to develop the learning components of adaptive 
information filtering agents, and discusses possible difficulties of
applying such a framework to these agents.
\end{quote}
\end{abstract}

\section{Introduction} 
\label{Intro}

With the explosive growth of the
Internet and the World Wide Web (\emph{Web}), it is
becoming increasingly difficult for users to retrieve relevant 
information. This is the so-called problem of \emph{information overload}
on the Internet.
Augmenting existing Internet search
tools with personalised information filtering agents is one possible
method to alleviate this problem.
Adaptive information filtering agents are computer systems situated on
the Web. They \emph{autonomously} filter the incoming stream of 
information on behalf of the users. 
Users' information needs will change over time, and
information filtering agents must be able to revise their beliefs
about the users' information needs so that
the accuracy of the filtering process can be maintained. 
The \textbf{AGM} belief revision paradigm~\cite{Article:85:Alchourron:Belief} 
provides a rich and rigorous foundation
for modelling such revision processes. It enables an agent to
modify its beliefs in a rational and minimal way.
\textbf{Maxi-adjustment}~\cite{Article:96:Wlliams:Belief,Article:97:Wlliams:Belief}
is a specific change strategy that follows the AGM's
rationale of belief revision. In particular, it transmutes the 
underlying entrenchment ranking of beliefs in an absolute minimal
way under maximal information inertia. 
In information retrieval models~\cite{Book:83:Salton:IntroIR,Book:89:Salton:AutomTextProc}, 
information objects are often assumed
independent unless semantic relationships among them can be derived.
This intuition coincides with the underlying
assumption of the maxi-adjustment strategy. 
The advantage of employing the maxi-adjustment strategy as the agents' learning
mechanism is that semantic relationships among information items
can be taken into account during the agents' learning and
adaptation processes. Less users' relevance feedback~\cite{Article:90:Salton:IR}
is required to train the filtering agents, and
hence a higher level of \emph{learning autonomy} can be achieved
when compared with the other learning approaches employed in adaptive
information 
agents~\cite{Article:99:Billssu:Agent,Article:98:Moukas:Agent,Article:97:Balabanovic:Agent,Article:96:Pazzani:Agent,Article:95:Armstrong:Agent}.
This paper focuses on the application of the maxi-adjustment method to
the development of learning mechanisms in adaptive information filtering agents. 
Moreover, difficulties of applying such a framework to the
filtering agents are discussed.

\section{The Adaptive Filtering Agent}
\label{Architecture}

\begin{figure}[!htb]
\centerline{\includegraphics[width=8cm,height=5cm]{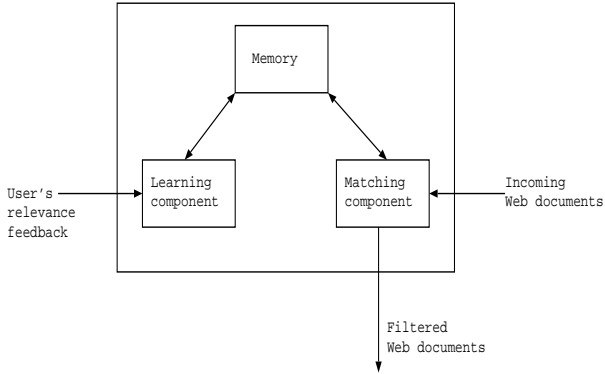}}
\caption{An Overview of Adaptive Information Filtering Agent}
\label{agent}
\end{figure}

Figure~\ref{agent}
is an overview of the major functional components of an adaptive
information filtering agent. 
The  focus  of  this   paper  is  on  the \emph{learning} component 
of the \emph{adaptive information filtering agent}.  
The filtering  agent's memory holds
the current representation of a user's  information
needs.  In particular, the notion of belief state~\cite{Book:88:Gardenfors:Belief}
is used to represent these information needs.
The learning component accepts a user's relevance 
feedback about filtered
Web documents. Based on the proposed induction method, these
feedback is converted to the beliefs of a user's information needs. The
maxi-adjustment strategy
is then applied to revise the beliefs stored in the agent's memory
with respect to these newly induced beliefs. As the
user's information needs will change over time, such a belief revision
process needs to be conducted repeatedly. 
Technically, the learning behavior demonstrated by the agent is a
kind of \emph{reinforcement learning}~\cite{Book:96:Langley:Learning}.    
The  matching component filters out relevant information from
a stream of incoming Web documents. It is
underpinned  by  logical deduction.  In other words, if the 
representation of a Web document is logically entailed by the
set of beliefs stored in the agent's memory, the Web document
will be considered as relevant and presented to the user. According to the
user's relevance feedback, new beliefs may be added to, or existing
beliefs may be contracted from the agent's memory. The adaptive
information filtering agent is one of the main elements in 
the agent-based information filtering system 
(AIFS)~\cite{Article:99:Lau:Agent}.

\section{The AGM Belief Revision Paradigm}
\label{BeliefRev}

The AGM paradigm~\cite{Article:85:Alchourron:Belief} provides a rigorous foundation for modelling 
consistent and minimal changes to an agent's beliefs. In particular, 
belief revision is taken as transitions among belief
states~\cite{Book:88:Gardenfors:Belief}. A belief state can be represented
by a belief set $K$ which is 
a theory in a propositional language $\plang$~\cite{Article:88:Gardenfors:Belief}. 
For the discussion in this paper, it is assumed that $\plang$ is the classical
propositional language. In the AGM framework~\cite{Article:85:Alchourron:Belief,Book:88:Gardenfors:Belief,Article:92:Gardenfors:Belief,Article:88:Gardenfors:Belief},
three principle types of belief state transitions are identified and 
modelled by corresponding belief functions: 
\emph{expansion} ($K^{+}_{\alpha}$), \emph{contraction} ($K^{-}_{\alpha}$),
and \emph{revision} ($K^{*}_{\alpha}$). 
The process of belief revision can be derived from the process of belief 
contraction and vice versa through the so-called 
\emph{Levi Identity} i.e.
$K^{*}_\alpha = (K^{-}_{\neg \alpha})^{+}_{\alpha} $, 
and the \emph{Harper Identity} i.e.
$K^{-}_\alpha = K \cap K^{*}_{\neg \alpha}$.
Essentially, the AGM framework proposes sets of
postulates to characterise these
functions such that they adhere to the
rationales of \emph{consistent} and \emph{minimal} changes. In addition,
it also describes the constructions of these
functions based on various mechanisms. One of them is  
\emph{epistemic entrenchment} 
($ \leqslant $)~\cite{Article:88:Gardenfors:Belief}. 
For instance, if $\alpha, \beta$ are beliefs in a belief set $K$, 
$\alpha \leqslant \beta$  means that $\beta$ is at least as entrenched
as $\alpha$.
If inconsistency arises after applying changes to a belief set,
beliefs with the lowest degree of epistemic entrenchment
are given up. Technically, epistemic 
entrenchment~\cite{Article:88:Gardenfors:Belief,Article:92:Gardenfors:Belief} 
is a total preorder of the sentences (e.g.\ $\alpha, \beta, \gamma$) in $\plang$,
and is characterised by the following postulates:  
\textbf{(EE1)}: If $\alpha \leqslant \beta$ and $\beta \leqslant \gamma$, then $\alpha \leqslant \gamma$;
\textbf{(EE2)}: If $\alpha \vdash \beta$, then $\alpha \leqslant \beta$;
\textbf{(EE3)}: For any $\alpha$ and $\beta$, $\alpha \leqslant \alpha \wedge \beta \textrm{ or } \beta \leqslant \alpha \wedge \beta $;
\textbf{(EE4)}: When $K \neq K_{\perp}, \  \alpha \notin K \textrm{ iff } \ \alpha \leqslant \beta$, for all $\beta$;
\textbf{(EE5)}: If $\beta \leqslant \alpha$ for all $\beta$, then $\vdash \alpha$.

It has been proved that an unique contraction function can be defined
by the underlying epistemic 
entrenchment through the (C-) condition~\cite{Article:88:Gardenfors:Belief}:
\begin{displaymath}
\textrm{(C-)} \qquad K^{-}_{\alpha} = \left\{ \begin{array}{ll}
		\{ \beta \in K: \alpha < \alpha \vee \beta \} & \textrm{if } \not \vdash \alpha \\
		K & \textrm{otherwise} \end{array} \right.
\end{displaymath}

where $<$ is the strict
part of epistemic entrenchment defined above. 
Moreover, the (C-R) condition~\cite{Article:92:Gardenfors:Belief} 
also ensures that if an ordering
of beliefs satisfies (EE1)-(EE5), the contraction function, uniquely 
determined by (C-R), satisfies all but the \emph{recovery} postulates for contraction.
\begin{displaymath}
\textrm{(C-R)} \qquad K^{-}_{\alpha} = \left\{ \begin{array}{ll}
		\{ \beta \in K: \alpha < \beta \} & \textrm{if } \not \vdash \alpha \\
		K & \textrm{otherwise} \end{array} \right.
\end{displaymath}

Nevertheless, for a computer based implementation, a finite
representation of epistemic entrenchment ordering 
and a policy of iterated belief changes are required.
Williams \cite{Article:95a:Williams:Belief,Chapter:97a:Williams:Belief} 
proposed the \emph{finite partial entrenchment ranking} $(\B)$ that 
ranked the sentences of a theory in $\plang$ with the minimum possible
degree of entrenchment $(\leqslant_{\B})$. Moreover, 
\emph{maxi-adjustment}~\cite{Article:96:Wlliams:Belief,Article:97:Wlliams:Belief} 
was proposed to transmute a finite partial ranking using an absolute measure 
of minimal change under maximal information inertia. Belief revision
is not just taken as adding
or deleting a sentence from a theory but the 
transmutation of the underlying entrenchment ranking. 
Williams~\cite{Article:95a:Williams:Belief,Article:96:Wlliams:Belief,Article:97:Wlliams:Belief}
formally defined the following definitions for a computational model of
belief revision.

\begin{definition} \label{rankB}
A finite partial entrenchment ranking is a function $\B$ that maps a finite
subset of sentences in $\plang$ into the interval $[0, \mathcal{O}]$ 
such that the following conditions are satisfied for all
$\alpha \in dom(\B)$:

(PER1) $\{ \beta \in \textrm{dom}(\B): \B(\alpha) < \B(\beta) \} \not \vdash \alpha$. 

(PER2) If $\vdash \neg \alpha$ then $\B(\alpha) = 0$.

(PER3) $\B(\alpha) = \mathcal{O}$ if and only if $\vdash \alpha$.
\end{definition}

The set of all partial entrenchment rankings is
denoted $\mathcal{B}$. $\B(\alpha)$ is referred as the 
\emph{degree of acceptance} of
$\alpha$. The explicit information content of $\B \in \mathcal{B}$ is 
$\{ \alpha \in \textrm{dom}(\B): \B(\alpha) > 0 \}$, and is denoted
$exp(\B)$. Similarly, the implicit information content represented by 
$\B \in \mathcal{B}$ is
$Cn(exp(\B))$, and is denoted $content(\B)$. $Cn$ is the classical consequence
operator. In order to describe
the epistemic entrenchment ordering $(\leqslant_{\B})$ generated from 
a finite partial entrenchment ranking $\B$, 
it is necessary to rank implicit sentences.

\begin{definition} \label{degree}
Let $\alpha$ be a non tautological sentence. Let $\B$ be a finite
partial entrenchment ranking. The degree of acceptance of
$\alpha$ is defined as: 

\begin{displaymath}
degree(\B, \alpha) = \left \{ \begin{array}{l}
          \textrm{largest j such that} \\
          \{\beta \in exp(\B): \B(\beta) \geq j \} \vdash \alpha \\
          \qquad \qquad \ \textrm{if } \alpha \in content(\B) \\
          0 \qquad \qquad \textrm{otherwise}
          \end{array} \right.
\end{displaymath}
       
\end{definition}

The maxi-adjustment strategy transmutes a partial entrenchment ranking $\B$ based
on the rationale of absolute minimal change under maximal information inertia. 
It is assumed that sentences in exp($\B$) (e.g.\ $\alpha, \beta$) are 
independent unless logical dependence exists between them. In particular,
$\alpha$ is defined as a \emph{reason} of $\beta$ if and only if
$degree(\B, \alpha \rightarrow \beta) > \B(\beta)$.

\begin{definition} \label{maxi}
Let $\B \in \mathcal{B}$ be finite. The range of $\B$ is enumerated 
in ascending order as
$j_{0}, j_{1}, j_{2}, \ldots, j_{\mathcal{O}}$. Let
$\alpha$ be a contingent sentence, $j_{m} = degree(\B, \alpha)$ and
$ 0 \leq i < \mathcal{O}$. Then the $(\alpha, i)$ maxi-adjustment of
$\B$ is $\B^{\star}(\alpha, i)$ defined by:

\begin{displaymath}
 \B^{\star}(\alpha, i) = \left \{ \begin{array}{ll}
         (\B^{-}(\alpha, i)) & \textrm{if } i \leq j_{m}\\
         (\B^{-}(\neg \alpha, 0))^{+}(\alpha, i) & \textrm{otherwise}
         \end{array} \right.
\end{displaymath}

where for all $\beta \in dom(\B)$, $\B^{-}(\alpha, i)$ is defined as follows:

1. For $\beta$ with $\B(\beta) > j_{m}, \  \B^{-}(\alpha, i)(\beta) = \B(\beta)$.

2. For $\beta$ with $i < \B(\beta) \leq j_{m}$, assuming that $\B^{-}(\alpha, i)(\beta)$
for $\beta$ is defined with $\B(\beta) \geq j_{m-k}$ for $k = -1,0,1,2,\ldots,n-1$, 
then for $\beta$ with $\B(\beta) = j_{m-n}$,

\begin{displaymath}
 \B^{-}(\alpha, i)(\beta) = \left \{ \begin{array}{l}
         i \quad \textrm{if } \alpha \vdash \beta \textrm{ or}\\
           \quad \alpha \not \vdash \beta \textrm{ and } \beta \in \Gamma\\
           \quad \textrm{where } \Gamma \textrm{ is a minimal subset of}\\
           \quad \{ \gamma: \B(\gamma) = j_{m-n} \} \textrm{ such that}\\
           \quad \{ \gamma: \B^{-}(\alpha, i)(\gamma) > j_{m-n} \} \cup \Gamma \vdash \alpha\\
         \B(\beta) \quad \textrm{otherwise}
         \end{array} \right.
\end{displaymath}   

3. For $\beta$ with $\B(\beta) \leq i \ , \B^{-}(\alpha, i)(\beta) = \B(\beta)$.

For all $\beta \in dom(\B) \cup \{ \alpha \}, \  \B^{+}(\alpha, i)$ 
is defined as follows:

\begin{displaymath}
 \B^{+}(\alpha, i)(\beta) = \left \{ \begin{array}{l}
         \B(\beta) \qquad \textrm{if } \B(\beta) > i\\
          i \qquad \qquad \textrm{if } \alpha \equiv \beta \textrm{ or } \\
           \qquad \B(\beta) \leq i < degree(\B, \alpha \rightarrow \beta)\\
          degree(\B, \alpha \rightarrow \beta) \qquad \textrm{otherwise}\\
           \end{array} \right.
\end{displaymath}   

\end{definition}

It has been stated that if $i > 0$ then
$content(\B^{\star}(\alpha, i))$ is an AGM revision, and
$content(\B^{\star}(\alpha, 0))$ satisfies all but the \emph{recovery} postulates
for AGM contraction~\cite{Article:96:Wlliams:Belief}.

\section{Knowledge Representation}
\label{Representation}

A Web page is characterised by a set of weighted keywords based on
traditional information retrieval (\emph{IR}) 
techniques~\cite{Book:83:Salton:IntroIR,Book:89:Salton:AutomTextProc}. 
At the symbolic level, each keyword $k$ is mapped to the ground term of 
the \emph{positive keyword} predicate \emph{pkw} i.e.\ $pkw(k)$.
Basically, $pkw(k)$ is
a \emph{proposition} since its interpretation is either true or false. 
The intended interpretation of these sentences is that they
are satisfied in a document $D$ i.e.\ $D \models pkw(k)$ if $D$ is 
taken as a model~\cite{Article:92:Chiaramella:IRLogic,Article:98:Lalmas:LogicIR}. 
For example, if d = \{ business, commerce, trade, \ldots \} is the document
representation at the keyword level, the corresponding 
representation at the symbolic level will be 
d = \{ $pkw(business)$, $pkw(commerce)$, $pkw(trade)$, \ldots \}.
Similarly, a user's information needs are also represented as a set
of weighted keywords at the keyword level. However, this set 
of weighted keywords is derived from a set of relevant documents $D^{+}$
and a set of non-relevant documents $D^{-}$
with respect to the user's information
needs~\cite{Article:96:Allan:IF,Article:94:Buckley:IF}. 
Based on the frequencies of these keywords appearing in
$D^{+}$ and $D^{-}$, it is possible to induce a preference ordering 
among the keywords with respect to the user's information needs. 
The basic idea is that a keyword appearing more frequently
in $D^{+}$ is a more preferred keyword than another keyword that appears
less frequently in $D^{+}$. Once this preference ordering is induced, it
is taken as the epistemic entrenchment ordering of the
corresponding beliefs. It is observed that the postulates of epistemic
entrenchment is valid in the context of information retrieval in 
general~\cite{Article:99:Lau:Agent}. For example, if $\{ \alpha, \beta, \gamma \}$
is a set of information carriers~\cite{Article:94:Bruza:Axioms}, $\alpha \leqslant \beta$ and $\beta \leqslant \gamma$
implies $\alpha \leqslant \gamma$. In other words,
if an information searcher prefers information carrier $\gamma$ rather than
information carrier $\beta$, and information carrier $\beta$ rather than
information carrier $\alpha$, they prefer retrieving $\gamma$ rather than 
$\alpha$. This characteristic of information
carriers matches the epistemic entrenchment postulate e.g. (EE1) of beliefs.

Moreover, it is necessary to classify a keyword  as \emph{positive}, 
\emph{neutral}, or \emph{negative}~\cite{Article:97:Kindo:IF}. 
Intuitively, positive keywords represent the information
items in which the users interested. Negative keywords represent
the information items that the users do not want to retrieve. Neutral
keywords mean that these keywords are not useful for determining  
the users' interests. 
Eq.(\ref{eq1}) is developed based on the \emph{keyword classifier}~\cite{Article:97:Kindo:IF}.
It can be used to induce the preference value $pre(k)$ of a keyword 
$k$, and classify it as positive, negative, or neutral.

\begin{equation} \label{eq1}
\begin{array}{l}
pre(k) = \epsilon \times \tanh \left( \frac{df(k)}{\xi} \right) \times \\
          \left( p(k_{rel}) \tanh \frac{ p(k_{rel}) }{p_{rel}} -
          ( 1 - p(k_{rel}) ) \tanh \frac{ (1 - p(k_{rel})) }{( 1 - p_{rel})}
          \right)\\
\end{array}        
\end{equation}

where $\epsilon$ is used to restrict the range of $pre(k)$ such 
that $-1 < pre(k) < 1$.
The examples illustrated in this paper assume that $\epsilon = 0.9$.
$df(k)$ is the sum of the number of relevant documents $df(k_{rel})$ and
the number of non-relevant documents $df(k_{nrel})$ that contains the 
keyword $k$, and $\tanh$ is the hyperbolic tangent. 
The rarity parameter $\xi$ is used to control rare or 
new keywords and is expressed as int$(logN + 1)$,
where $N$ is the total number of Web documents judged by a user, and
int is an integer function that truncates the decimal values. 
$p(k_{rel})$ is the estimated probability that
a document containing keyword $k$ is relevant and is expressed as the
fraction $\frac{df(k_{rel})}{df(k_{rel}) + df(k_{nrel})}$. $p_{rel}$ is
the estimated probability that a document is relevant. 
In our system, it is assumed that 
the probability that a Web document presented by the filtering
agent and judged as relevant by a user is $p_{rel} = 0.5$. 
A positive value of $pre(k)$ implies that the associated keyword is positive,
whereas a negative value of $pre(k)$ indicates a negative 
keyword. If $pre(k)$ is below a threshold 
value $\lambda$, the associated keyword is considered neutral. 
It is assumed that $\lambda = 0.5$ for the examples demonstrated in this
paper. Basically a positive keyword $k$ is mapped to $pkw(k)$, 
and a negative keyword $k$ is mapped to $\neg pkw(k)$.
There is no need to create the symbolic representations for neutral keywords.
For $pkw(k)$ or $\neg pkw(k)$, the 
entrenchment rank $\B(\alpha_{k})$ of the corresponding formula $\alpha_{k}$ 
is defined as:

\begin{equation} \label{eq2}
\B(\alpha_{k}) = \left \{ \begin{array}{ll}
                   |pre(k)| & \textrm{if } |pre(k)| \geq \lambda \\
                    0 & \textrm{otherwise}
                    \end{array} \right.   
\end{equation}

The following is an example
of computing the entrenchment rank $\B(\alpha_{k})$ from a set of judged Web documents. 
It is assumed that there are
a set of five documents having been judged as relevant (i.e. $|D^{+}| = 5$)
and another set of five documents having been judged as non-relevant 
by a user (i.e. $|D^{-}| = 5$). Each document is characterised by a set of 
keywords e.g.\ $d_{1}=\{business, commerce, system \}, d_{2}=\{art, sculpture \}$.
Table~\ref{tab:frequency} summarises the frequencies of these keywords 
appearing in both $D^{+}$ and $D^{-}$, their preference values, and the
entrenchment ranks of corresponding formulae.

\begin{table}[!htb]
\begin{center}
\caption{ \small Representation of users' information preferences} \label{tab:frequency}
\begin{tabular}{|c|c|c|c|c|c|}
  \hline
  \tiny Keywords & \tiny $D^{+}$ & \tiny $D^{-}$ & \tiny $pre(k)$ & \tiny Formula:$\alpha_{k}$ & \tiny Rank:$\B(\alpha_{k})$ \\
  \hline
  \tiny business & \tiny 5 & \tiny 0 & \tiny 0.856 & \tiny $pkw(business)$ & \tiny 0.856\\
  \tiny commerce & \tiny 4 & \tiny 0 & \tiny 0.836 & \tiny $pkw(commerce)$ & \tiny 0.836\\
  \tiny system & \tiny 2 & \tiny 2 & \tiny 0 & \tiny - & \tiny -\\
  \tiny art & \tiny 0 & \tiny 5 & \tiny -0.856 & \tiny $\neg pkw(art)$ & \tiny 0.856\\
  \tiny sculpture & \tiny 0 & \tiny 3 & \tiny -0.785 & \tiny $\neg pkw(sculpture)$ & \tiny 0.785\\
  \tiny insurance & \tiny 1 & \tiny 0 & \tiny 0.401 & \tiny - & \tiny -\\
  \hline
\end{tabular}
\end{center}
\end{table}

\section{Learning and Adaptation}
\label{Learning}

Whenever a user provides relevance feedback for a presented Web document,
the belief revision process can be invoked to learn the user's
current information preferences.
Conceptually, the filtering agent's learning and adaptation mechanism 
is characterised by
the belief \emph{revision} and \emph{contraction} processes. 
For example, if $\Gamma = \{a_{1}, a_{2}, \ldots, a_{n} \}$
is a set of formulae representing a Web document $d \in D^{+}$,
the belief revision process 
$(((K^{*}_{a_{1}})^{*}_{a_{2}}) \ldots )^{*}_{a_{n}}$ 
is invoked for each $a_{i} \in \Gamma$, where $K$ is the belief set stored
in the filtering agent's memory.
On the other hand, the belief contraction process 
$(((K^{-}_{a_{1}})^{-}_{a_{2}}) \ldots )^{-}_{a_{n}}$
is applied for each $a_{i} \in \Gamma$ and $d \in D^{-}$.
The sequence of revising or contracting the set of beliefs $a_{1}, a_{2}, \ldots, a_{n}$ is
determined by their entrenchment ranks and whether it is a revision or contraction
operation.
At the computational level, belief revision is actually taken as the adjustment of entrenchment ranking
$\B$ in the \emph{theory base} $exp(\B)$.  Particularly, 
\emph{maxi-adjustment} \cite{Article:96:Wlliams:Belief,Article:98b:Williams:Belief}
is employed by the learning component of the filtering agent to modify the ranking 
of its beliefs in an absolute minimal way under maximal information
inertia. 
As the input to the maxi-adjustment
algorithm consists of a sentence $\alpha$ and its entrenchment rank $i$, 
the procedure described in the Knowledge Representation
Section is used to induce the new rank for each $\alpha \in \Gamma$.
Moreover, for our implementation of the maxi-adjustment algorithm, the
maximal ordinal $\mathcal{O}$ in the interval $[0,\mathcal{O}]$ is chosen as $1$.

One advantage of a symbolic representation of
the filtering agent's domain knowledge is that semantic
relationships among keywords can be captured.  
For example, if the keywords \emph{business}
and \emph{commerce} are taken as synonymous, this semantic relationship
can be modelled as a formula $pkw(business) \equiv pkw(commerce)$ in 
the filtering agent's memory. Moreover, classification knowledge such
as \emph{sculpture} is a kind of \emph{art} can also be used by
specifying the rule such
as $pkw(sculpture) \rightarrow pkw(art)$. It is 
believed that capturing the semantic relationships among keywords 
can improve the effectiveness of the matching 
process~\cite{Article:95:Hunter:DefaultLogic,Article:95:Nie:IRLogic}. In fact, 
by employing maxi-adjustment as the filtering agent's learning mechanism,
these semantic relationships can be reasoned about  
during the reinforcement learning process. 
This could lead to a higher level of \emph{learning autonomy} since
changes to related keywords can automatically be inferred by the filtering agent. 
As a result, less users' relevance feedback may be required.
The following examples assume that the formulae 
$pkw(business)$ $\equiv$ $pkw(commerce)$ and
$pkw(sculpture)$ $\rightarrow$ $pkw(art)$ have been manually added to the
the agent's memory through a knowledge engineering process.\\

\textbf{Example 1:}

The first example shows how adding one belief to the agent's memory
will automatically raise the entrenchment rank of another related belief. 
It is assumed that the belief $\B(\neg pkw(sculpture)) = 0.785$ 
and the belief $\B(pkw(business)) = 0.856$ have been
learnt by the filtering agent.
If several Web documents characterised by the keyword
\emph{art} are also judged as non-relevant by the user later on, the preference
value of the keyword \emph{art} can be induced according to Eq.(\ref{eq1}).
Assuming that $pre(art) = -0.856$, the corresponding entrenchment rank can 
be computed as $\B(\neg pkw(art)) = 0.856$ according to Eq.(\ref{eq2}). 
By applying $\B^{\star}(\neg pkw(art), 0.856)$ to the theory base
$exp(\B)$, the before and after images of the agent's \emph{explicit beliefs} 
(i.e. $exp(\B)$) can be
tabulated in Table~\ref{tab:ex1}. Based on the maxi-adjustment 
algorithm, $\B^{+}(\alpha, i)(\beta) = i$ if
$\B(\beta) \leq i < degree(\B, \alpha \rightarrow \beta)$.

\begin{displaymath}
\begin{array}{l}
\because \B(\neg pkw(sculpture)) \leq 0.856 < \\
\quad degree(\B, \neg pkw(art) \rightarrow \neg pkw(sculpture))\\
\therefore \B^{+}(\neg pkw(art), 0.856)(\neg pkw(sculpture)) = 0.856\\
\end{array}
\end{displaymath}

The implicit belief $\neg pkw(art)$ $\rightarrow$ $\neg pkw(sculpture)$
in $content(\B)$ is derived from the explicit belief $pkw(sculpture)$ $\rightarrow$ $pkw(art)$
in the theory base $exp(\B)$, and its degree of acceptance is $1$
according to Definition~\ref{degree}.
As the belief $\neg pkw(art)$ implies the belief $\neg pkw(sculpture)$ and
the agent believes in $\neg pkw(art)$, the
belief $\neg pkw(sculpture)$ should be at least as entrenched as the
belief $\neg pkw(art)$ according to (PER1) of Definition~\ref{rankB} or
(EE2). In other words, whenever the agent believes that the user is
not interested in \emph{art} (i.e. $\neg pkw(art)$),
it must be prepared to accept that the user is also not interested in
\emph{sculpture} at least to the degree of the former. 
The proposed learning and
adaptation framework is more effective than other learning approaches that
can not take into account the semantic relationships among information
items. This example demonstrates the automatic revision of the agent's
beliefs about related keywords given the relevance feedback for a 
particular keyword. Therefore, less users' relevance feedback may be 
required during reinforcement learning. Consequently, \emph{learning autonomy}
of the filtering agent can be enhanced.

\begin{table}[!htb]
\begin{center}
\caption{\small Raising related beliefs} \label{tab:ex1}
\begin{tabular}{|c|c|c|}
  \hline 
  \tiny Formula:$\alpha$ & \tiny $\B(\alpha)$ Before &  \tiny $\B(\alpha)$ After  \\
  \hline
  \tiny $pkw(business) \equiv pkw(commerce)$ & \tiny 1.000 & \tiny 1.000\\
  
  \tiny $pkw(sculpture) \rightarrow pkw(art)$ & \tiny 1.000 & \tiny 1.000\\
  \tiny $pkw(business)$ & \tiny 0.856 & \tiny 0.856\\
  \tiny $\neg pkw(sculpture)$  & \tiny 0.785 & \tiny \textbf{0.856}\\
  \tiny $\neg pkw(art)$  & \tiny 0 & \tiny \textbf{0.856}\\
  \hline
\end{tabular}
\end{center}
\end{table}

\textbf{Example 2:}

The second example illustrates the belief contraction process. In particular,
how the contraction of one belief will automatically remove
another related belief from the agent's memory if there is a semantic 
relationship between the underlying keywords. 
Assuming that more Web documents characterised by the keyword \emph{sculpture} are 
judged as relevant by the user at a later stage, the belief $\B(pkw(sculpture)) = 0.785$
could be induced. 
As $\B^{\star}(\alpha, i)=(\B^{-}(\neg \alpha, 0))^{+}(\alpha, i)$
if $i > j_{m}$, where $i = 0.785$ and $j_{m} = 0$ in this example,
$\B^{\star}($ $pkw(sculpture),$ $0.785)$ leads to the contraction of the
belief $\neg pkw(sculpture)$ from the theory base $exp(\B)$. 
Moreover, $\B^{-}($ $\neg$ $pkw(sculpture),$ $0)$
$(\neg$ $pkw(art))$ $=$ $0$ is computed
because $\B($ $\neg$ $pkw($ $art))$ $=$ $0.856$ $=$ $j_{m-n}$ and the set
$\{\gamma:$ $\B^{-}($ $\neg$ $pkw(sculpture),$ $0)$ $(\gamma)$ $>$ $0.856$ $\}$ 
$\cup$ $\{$ $\neg$ $pkw(art)$ $\}$ $\vdash$ $\neg$ $pkw(sculpture)$ is obtained.
The before and after images of the filtering agent's explicit beliefs
are tabulated in Table~\ref{tab:ex2}.

\begin{table}[!htb]
\begin{center}
\caption{\small Contracting related beliefs} \label{tab:ex2}
\begin{tabular}{|c|c|c|}
  \hline 
  \tiny Formula:$\alpha$ & \tiny $\B(\alpha)$ Before &  \tiny $\B(\alpha)$ After  \\
  \hline
  \tiny $pkw(business) \equiv pkw(commerce)$ & \tiny 1.000 & \tiny 1.000\\
  
  \tiny $pkw(sculpture) \rightarrow pkw(art)$ & \tiny 1.000 & \tiny 1.000\\
  \tiny $pkw(business)$ & \tiny 0.856 & \tiny  0.856\\
  \tiny $pkw(sculpture)$ & \tiny 0 & \tiny \textbf{0.785}\\
  
  \tiny $\neg pkw(sculpture)$  & \tiny 0.856 & \tiny \textbf{0}\\
  \tiny $\neg pkw(art)$  & \tiny 0.856 & \tiny \textbf{0}\\
  \hline
\end{tabular}
\end{center}
\end{table}

\textbf{Example 3:}

This example demonstrates how multiple sentences from the same
Web document judged by a user can be contracted from the agent's
memory. If some Web documents characterised by \emph{sculpture} and 
\emph{business} have recently been 
judged as non-relevant by the user, both the belief $\B(pkw(sculpture)) = 0$
and the belief $\B(pkw(business)) = 0$ could be induced. 
Since beliefs with minimal entrenchment rank
(i.e. $0$) are not supposed to be stored in the agent's memory, 
maxi-adjustement for these two beliefs is still required so that they can  
be removed from the agent's memory.
Basically, the contraction process $\B^{-}(\alpha, i)$ is applied to both
sentences. As opposed to the belief expansion process where the most entrenched
sentence is applied to the agent's memory first, the least entrenched
belief is first contracted from the agent's memory for belief contraction. 
Consequently, $\B^{-}(pkw(sculpture), 0)$ is invoked first.
As there is not other logically related sentences in the theory base $exp(\B)$,
the belief $pkw(sculpture)$ is simply removed from the agent's memory. 
Similarly, the sentence  $pkw(business)$ is also removed from $exp(\B)$
by applying $\B^{-}(pkw(business), 0)$.
The before and after images of the filtering agent's explicit beliefs
are tabulated in Table~\ref{tab:ex3}.

\begin{table}[!htb]
\begin{center}
\caption{\small Contracting multiple beliefs in one cycle} \label{tab:ex3}
\begin{tabular}{|c|c|c|}
  \hline 
  \tiny Formula:$\alpha$ & \tiny $\B(\alpha)$ Before &  \tiny $\B(\alpha)$ After  \\
  \hline
  \tiny $pkw(business) \equiv pkw(commerce)$ & \tiny 1.000 & \tiny 1.000\\
  
  \tiny $pkw(sculpture) \rightarrow pkw(art)$ & \tiny 1.000 & \tiny 1.000\\
  \tiny $pkw(business)$ & \tiny 0.856 & \tiny  \textbf{0}\\
  \tiny $pkw(sculpture)$ & \tiny 0.785 & \tiny \textbf{0}\\
  
  \hline
\end{tabular}
\end{center}
\end{table}

\section{Filtering Web Documents}
\label{Matching} 
 
In our current framework, the matching function of the filtering agent is modelled 
as logical deduction. Moreover, a Web document is taken as
the conjunction of a set of formulae~\cite{Article:92:Chiaramella:IRLogic,Article:95:Hunter:DefaultLogic}.
The following example illustrates the agent's
deduction process with reference to previous examples. 
The resulting belief sets $K=Cn(exp(\B))$ 
from examples $1$, $2$, and $3$ in the previous section are used to determine 
the relevance of the following three Web documents:

\begin{displaymath}
\begin{array}{l}
\phi = \{pkw(business) \wedge pkw(art)\} \\
\varphi = \{pkw(sculpture) \wedge pkw(art)\} \\
\psi = \{pkw(business) \wedge pkw(commerce)\}
\end{array}
\end{displaymath}

The filtering agent's conclusions about the relevance of the Web documents
are summarised as follows:\\

Time: (t1)
\begin{displaymath}
\begin{array}{l}
K = Cn(\{pkw(business) \equiv pkw(commerce),\\
\qquad pkw(sculpture) \rightarrow pkw(art),\\ 
\qquad pkw(business), \neg pkw(sculpture), \neg pkw(art) \})\\
\therefore \qquad K \not \vdash \phi, \quad K \not \vdash \varphi, \quad K \vdash \psi
\end{array}
\end{displaymath}

Time: (t2)
\begin{displaymath}
\begin{array}{l}
K = Cn(\{pkw(business) \equiv pkw(commerce),\\
\qquad pkw(sculpture) \rightarrow pkw(art),\\ 
\qquad pkw(business), pkw(sculpture) \})\\
\therefore \qquad K \vdash \phi, \quad K \vdash \varphi, \quad K \vdash \psi
\end{array}
\end{displaymath}

Time: (t3)
\begin{displaymath}
\begin{array}{l}
K = Cn(\{pkw(business) \equiv pkw(commerce),\\
\qquad pkw(sculpture) \rightarrow pkw(art) \})\\
\therefore \qquad K \not \vdash \phi, \quad K \not \vdash \varphi, \quad K \not \vdash \psi
\end{array}
\end{displaymath}

As can be seen tentative conclusion drawn at time $t1$ may not hold 
when new information e.g. $K^{-}_{pkw(business)}$ is processed by the agent
at time $t3$. Strictly
speaking, the deduction process of the agent should be described as
$K \nonmono \psi$, where $\nonmono$ is a 
nonmonotonic inference relation because the inferred beliefs (i.e. conclusions)
will not grow monotonically. It is not
difficult to see that this $\nonmono$ should belong to the class of
nonmonotonic inference called \emph{expectation inference} 
($\nonexpt$)~\cite{Article:94:Gardenfors:NMR}.
The basic idea of expectation inference is that given a sentence $\alpha$ of
a propositional language $\plang$, if $\alpha$ and the subset of sentences
in a belief set $K$ that is consistent with $\alpha$ 
(i.e. $\{\alpha \} \cup \{\beta \in K: \neg \alpha < \beta \}$) can
classically entail another sentence $\beta$, $ \alpha \nonexpt \beta $ can
be deduced. With reference to our examples, since $K$ classically entails
or not entails $\phi$, $\varphi$ and $\psi$, and $K \cup K = K$, 
the definition of expectation inference can trivially be applied 
to describe  the characteristics of the inference mechanism in 
the filtering agent. Therefore, 
$K \nonexpt \phi$, where $K$ is the filtering agent's belief set and $\phi$
is the logical representation of a Web document, represents an inference
conducted by the adaptive filtering agent.

\section{Discussion}
\label{Discussion}

An alternative approach for developing the filtering agent's
learning mechanism is to apply the (C-) or the (C-R) 
condition to revise the agent's beliefs. It has been stated that both the (C-)
and the (C-R) conditions can be used to construct the same class of belief
revision functions that satisfy the AGM postulates for belief
revision~\cite{Article:92:Gardenfors:Belief}. The following example illustrates
how belief revision may be conducted based on the (C-R) condition. Assuming
that the filtering agent's initial belief set $K$ is as follows:

\begin{displaymath}
\begin{array}{l}
K = \{pkw(business), pkw(commerce), \\
\quad pkw(sculpture), pkw(art) \}\\
\end{array}
\end{displaymath}

Moreover,

\begin{displaymath}
\begin{array}{l}
pkw(business) \geqslant pkw(commerce) \geqslant \\
\quad pkw(sculpture) \geqslant pkw(art)\\
\end{array}
\end{displaymath}

If a user perceives some Web documents characterised by the keyword \emph{business}
as non-relevant, the revision process $K^{*}_{\neg pkw(business)}$
will be invoked. Based on the Levi Identity, the sentence $pkw(business)$ should
first be contracted from the belief set $K$. Consequently, all
the beliefs from $K$ will be contracted according to the (C-R) condition.
The resulting belief set becomes $K^{'} = \{ \neg pkw(business) \}$.
Nevertheless, a user who does not require information objects about
\emph{business} may still be interested in information objects about
\emph{art} and \emph{sculpture}. Therefore, applying (C-R) or (C-) to
construct belief contraction or belief revision function seems
producing drastic changes in the context of information retrieval and filtering.
In fact, term independence is often assumed in information retrieval.
This intuition is reflected in the vector space 
model of information retrieval~\cite{Book:83:Salton:IntroIR}, where 
changing the weight of a particular keyword may not affect the others 
in the weight vector.
Therefore, the maxi-adjustment strategy produces a better approximation
in terms of revising or contracting beliefs about information objects
with respect to a user's information needs.

Under the current framework, domain knowledge such as semantic relationships
among information objects is transferred to the filtering agent's
memory through a knowledge engineering process. Since not all
semantic relationship is highly certain (i.e. assigning the maximal
entrenchment rank), by applying 
belief revision to the agent's memory, the corresponding beliefs
may be contracted from the memory over time. Therefore, domain knowledge
perhaps needs to be transferred to the agent's memory periodically in
accordance with the postulates of
epistemic entrenchment. This can be taken as an off-line process to 
minimise its impact on the availability of the filtering agents. 
However, further investigation is required to apply such a
background learning process to the filtering agents. 

The belief set $K = content(\B)$ is actually used by the agents to infer 
the relevance of Web documents. 
On the other hand, maxi-adjustment is employed to revise the theory base 
$exp(\B)$ and to maintain its consistency after applying changes. 
Though maxi-adjustment ensures
that the revised theory base is consistent, it is possible that
the belief set $K$ becomes inconsistent  (i.e. $K  \ \vdash  \ \perp$) 
after applying changes such as $\B^{\star}(\alpha, i)$. 
The following is a classical example to explain such a problem. 
It is assumed that the set of explicit beliefs $exp(\B)$ as well as their 
entrenchment ranking is as follows:

\begin{displaymath}
\begin{array}{l}
exp(\B) = \{ \forall x [penguin(x) \rightarrow bird(x)],\\
\qquad  \forall x [penguin(x) \rightarrow \neg fly(x)], \\
\qquad  \forall x [bird(x) \rightarrow fly(x)] \} \\
\end{array}
\end{displaymath}

\begin{displaymath}
\begin{array}{l}
\B(\forall x [penguin(x) \rightarrow bird(x)]) = 0.9,\\
\B(\forall x [penguin(x) \rightarrow \neg fly(x)]) = 0.7, \\
\B(\forall x [bird(x) \rightarrow  fly(x)]) = 0.4 \\
\end{array}
\end{displaymath}

In the context of information retrieval and filtering, $penguin(x)$ 
$\rightarrow$ $bird(x)$ can be interpreted as: if a user is interested in
information objects about $penguin$, it is likely that the user is also
interested in information objects about $bird$. The other formulae in
$exp(\B)$ can be interpreted in similar way. Given the fact
that the user is interested in $tweety$ which is a penguin 
i.e.\ $penguin(tweety)$ and $\B(penguin(tweety)) = 0.8$, by applying
maxi-adjustment $\B^{\star}(penguin(tweety), 0.8)$, the revised
entrenchment ranking $new\B$ will be:

\begin{displaymath}
\begin{array}{l}
new\B(\forall x [penguin(x) \rightarrow bird(x)]) = 0.9, \\
new\B(penguin(tweety)) = 0.8, \\
new\B(\forall x [penguin(x) \rightarrow \neg fly(x)]) = 0.7, \\
new\B(\forall x [bird(x) \rightarrow  fly(x)]) = 0.4\\
\end{array}
\end{displaymath}

The degree of acceptance of implicit beliefs is computed based on 
definition~\ref{degree}:

\begin{displaymath}
\begin{array}{l}
degree(new\B, \neg fly(tweety)) = 0.7, \\
degree(new\B, fly(tweety)) = 0.4\\
\end{array}
\end{displaymath}

As can be seen, even though $exp(new\B) \not \vdash \ \perp$, it is clear
that $content(new\B) \vdash \ \perp$, where $\vdash$ is classical derivability
relation. If the filtering agent employs 
the belief set $K = content(new\B)$
to deduce the relevance of Web documents, any documents will be
considered as relevant.
This problem must be addressed before the
filter agents can be  put to practical use. 
One possible solution is to make
use of the degree of acceptance of beliefs to produce the largest \emph{cut} of 
$content(\B)$ so that it does not entail $\perp$. 
In other words, only the set of consistent beliefs 
$\Delta = \{ \beta \in K: \ \perp \ < \beta \}$, where $<$ is the strict part
of epistemic entrenchment, will be used by the filtering agent
to infer the relevance of Web documents.
Similar idea has been explored in developing the expectation inference
relation~\cite{Article:94:Gardenfors:NMR}.
For instance,
$\alpha \nonexpt \gamma$ iff $\gamma \in Cn(\{\alpha\} \cup \{ \beta \in \Delta: \neg \alpha < \beta \})$
is developed based on $\Delta = \{ \beta: \ \perp \ < \beta \}$.
So, with reference to  the above example, after applying $\B^{\star}(penguin(tweety), 0.8)$, the agent 
should only make use of the following set of beliefs for reasoning: 

\begin{displaymath}
\begin{array}{l}
\Delta = \{ \forall x [penguin(x) \rightarrow bird(x)],\\
\qquad penguin(tweety), \\
\qquad bird(tweety), \\
\qquad  \forall x [penguin(x) \rightarrow \neg fly(x)], \\
\qquad \neg fly(tweety) \} \\
\end{array}
\end{displaymath}

Therefore, the agent can conclude
that the user is interested in information objects about 
\emph{non-flying tweety}. The above reasoning process can also
be explained based on \emph{nontrivial possibilistic deduction} 
($\nonP$)~\cite{Book:93:Dubois:Possibility,Article:94:Dubois:Possibility}.
In possibilistic logic, 
the inconsistency degree $Incons(F)$ of a possibilistic knowledge base $F$
is defined as the least certain formula involved in the strongest contradiction
of $F$. Moreover, nontrivial possibilistic deduction is defined as: 
$F \nonP (\alpha, m) \textrm{ iff } F \models (\alpha, m) \textrm{ and } m > Incons(F)$. 
If the entrenchment rank of a formula $\alpha$ is taken as the certainty $m$ of a
possibilistic formula, $Incons(K) = 0.4$ with reference to the above example.
By employing possibilistic resolution, $K \models (fly(tweety), 0.4)$ and
$K \models (\neg fly(tweety), 0.7)$ can be obtained, where $\models$ is 
possibilistic entailment. 
Since the certainty degree $m$ of $\neg fly(tweety)$ equals $0.7$ and is 
greater than $Incons(K)$, $K \nonP \neg fly(tweety)$. Nevertheless, 
$fly(tweety)$ can not be deduced from $K$ based on ($\nonP$) because
$m = 0.4 \not > Incons(K)$.
However, further investigation is required to apply possibilistic
based inference to the matching components of adaptive information
filtering agents.

\section{Conclusions}
\label{Conclusion}

The AGM belief revision paradigm offers a powerful and rigorous foundation
to model the changes of an agent's beliefs. 
The maxi-adjustment strategy, which follows the AGM rationale of consistent 
and minimal belief changes, provides a robust and effective computational mechanism for 
the development of the filtering agents' learning components.
As semantic relationships among information items can be reasoned about via the
maxi-adjustment method, less human intervention may be required
during the agents' reinforcement learning processes. 
This opens the door to better learning autonomy 
in adaptive information filtering agents. 
The technical feasibility of applying the maxi-adjustment method to
adaptive information filtering agents has been examined.
However, quantitative evaluation of the effectiveness of these agents needs
to be conducted to verify the advantages of applying such a framework to 
construct the learning mechanisms of these agents.

\subsubsection*{Acknowledgments}
The work reported in this paper has been funded in part by the
Cooperative Research Centres Program through the Department of
the Prime Minister and Cabinet of Australia.


\end{document}